\documentclass[conference]{IEEEtran}
\usepackage{times}

\usepackage{multicol}
\usepackage{graphicx}
\usepackage{dblfloatfix}
\usepackage[ruled]{algorithm2e}

\usepackage[numbers]{natbib}

\usepackage[ruled]{algorithm2e}

\SetCommentSty{mycommfont}
\SetFuncSty{myprocname}

\usepackage{amsmath, amsthm}
\usepackage{bm}
\usepackage{bbm}
\usepackage{amssymb}
\usepackage{upgreek}
\usepackage{multirow}
\usepackage{amsfonts}
\usepackage{gensymb}
\usepackage{here}
\usepackage[normalem]{ulem}
\usepackage{booktabs}
\usepackage{color}
\usepackage{subcaption}
\usepackage{url}
\usepackage[font=small, skip=5pt]{caption}
\usepackage{enumerate}
\usepackage{listings}
\usepackage[bookmarks=true]{hyperref}
\usepackage{xcolor} 
\hypersetup{
    colorlinks,
    linkcolor={red!50!black},
    citecolor={blue!80!black},
    urlcolor={blue!80!black}
}

\usepackage{cleveref}

\makeatletter
\def\@IEEEsectpunct{.\ \,}
\def\paragraph{\@startsection{paragraph}{4}{\z@}{1.5ex plus 1.5ex minus 0.5ex}%
{0ex}{\normalfont\normalsize\bfseries}}
\makeatother

\graphicspath{{figs/}}

\definecolor{purple}{rgb}{0.858, 0.08, 0.85}
\definecolor{darkgreen}{rgb}{0.08, 0.55, 0.08}
\definecolor{blue1}{rgb}{0.2, 0.2, 0.6}
\definecolor{green1}{rgb}{0.2, 0.6, 0.2}
\definecolor{green2}{rgb}{0.1, 0.4, 0.1}
\definecolor{dkgreen}{rgb}{0,0.6,0}
\definecolor{gray}{rgb}{0.5,0.5,0.5}
\definecolor{mauve}{rgb}{0.58,0,0.82}

\begin{document}

\title{Dialogue Object Search}

\author{
  Monica Roy\textsuperscript{*},
  Kaiyu Zheng\textsuperscript{*},
  Jason Liu,
  Stefanie Tellex\\
  Department of Computer Science, Brown University
}

\maketitle
\begingroup\renewcommand\thefootnote{*}
\footnotetext{These authors contributed equally to this work.}
\endgroup

\begin{abstract}
  We envision robots that can collaborate and communicate seamlessly with
  humans.  It is necessary for such robots to decide both what to say and how to
  act, while interacting with humans.  To this end, we introduce a new task,
  \emph{dialogue object search}: A robot is tasked to search for a target object
  (e.g. fork) in a human environment (e.g., kitchen), while engaging in a ``video
  call'' with a remote human who has additional but inexact knowledge about the
  target's location. That is, the robot conducts speech-based dialogue with the human,
  while sharing the image from its mounted camera. This task is challenging at
  multiple levels, from data collection, algorithm and system development, to
  evaluation. Despite these challenges, we believe such a task blocks the path
  towards more intelligent and collaborative robots.  In this extended abstract,
  we motivate and introduce the dialogue object search task and analyze examples
  collected from a pilot study. We then discuss our next steps and conclude with
  several challenges on which we hope to receive feedback.
\end{abstract}

\IEEEpeerreviewmaketitle

\section{Introduction}
Humans can act in the physical world (such as walking, looking, or opening a cabinet) while having a conversation with others. As robots enter homes and care centers, we envision them to have such capability as well when collaborating and communicating with humans. To achieve this, robots must decide both what to say and how to act towards a goal. This involves combining task-oriented dialogue systems with decision making under uncertainty for embodied agents. Traditionally, dialogue systems have involved users interacting with a virtual agent for tasks such as technical support \cite{mouromtsev2015spoken}, personal assistance (e.g., Siri) and booking reservations~\cite{wen2016network,wei2018airdialogue}. While recent works have proposed datasets that combine dialogue and dynamic, embodied decision making \cite{de2018talk,thomason2020vision}, the investigated problems over these datasets are limited to prediction tasks that bypass the challenges of evaluating a conversational embodied agent. For example, the Navigation from Dialog History Task \cite{thomason2020vision} asks the agent to predict the next navigation action, given a history of dialogue and past navigation actions. The tourist localization task \cite{de2018talk} asks the system to predict a location given a language description.

Our goal is to enable robots to naturally engage in a dialogue with a human
while completing a task autonomously. We believe a task that captures the
sequential nature of both the dialogue and physical decision making is necessary
for in-depth study towards this goal. We choose to focus on object search, a
useful and widely-studied problem~\cite{aydemir2013avo,kollar2009utilizing,zheng2020spatial,zheng2020multi}, and introduce a new task: \emph{dialogue object search}. Before providing a detailed description in the next section, we note that we consider speech-based dialogue in this task.
From the pilot study (Sec.~\ref{sec:pilot}), we observed that participants produced language and behavior that are more natural using speech, because text-based dialogue requires users to decide whether to type or act at every step.
Although this creates more challenges in scalable data collection and evaluation, we believe that overcoming these challenges is essential towards our goal, and they are our ongoing focus.

\section{Dialogue Object Search}
\begin{figure*}[tb]
    \centering
    \includegraphics[width=0.97\linewidth]{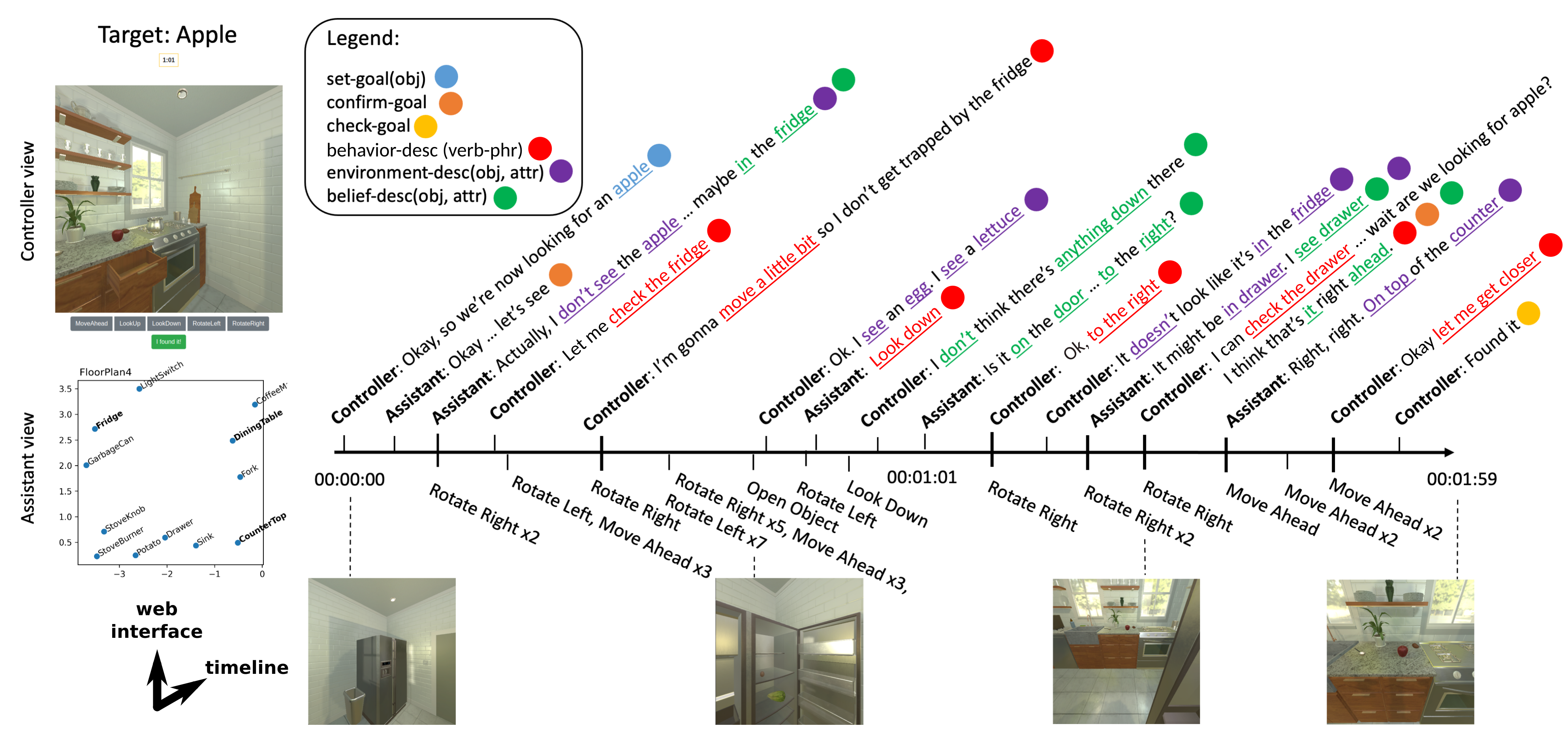}
    \caption{We conducted a pilot study to understand desirable behavior for the dialogue object search task. Shown here is a screenshot of the web interface (left) and the dialogue and actions organized onto a timeline (right), for an object search trial where the target object is \texttt{Apple}. We classified the dialogue utterances into a preliminary set of parameterized intents, indicated by the colors.}
    \label{fig:timeline}
    \vspace{-0.5cm}
\end{figure*}

A robot is tasked to search for a target object in a human
environment (e.g., kitchen) while engaging in an audio
dialogue with a remote human assistant, who possesses inexact prior knowledge
about the target object's location. In our pilot study, this is given in the form of a 2D scatter plot (Fig.~\ref{fig:timeline}). The robot
has a mounted RGB-D camera, and shares its view with the human assistant. We
assume the robot and the human assistant have access to two different sequences
of RGB-D images of the scene, which represent their prior experiences of living in
that environment. Target objects are excluded from these images.
The robot must decide what to say and how to act, in
order to efficiently find the target object while naturally interacting and
collaborating with the human assistant.


Our inspiration for the above setting comes from the following scenario
between two people living together (family or friends).  One person is searching
for something, such as a document or a key, but not sure where it is. They
decide to video call the other home member who is currently out of the house but may have a better idea.  They then engage in a dialogue while the first person conducts the search for the target object.
We envision that in the future, this could happen between a home assistant robot and a human user.

\section{Pilot Study}
\label{sec:pilot}

To investigate the above task, we first attempted to understand how a human would behave if they are in the robot's position. We designed and conducted a pilot study among three pairs of people (authors' lab members) using AI2-THOR~\cite{kolve2017ai2} as the simulated home environments. In this study, we designate two roles according to the above problem setting. The \emph{Assistant} is the person assisting in the process as the robot searches for a given target object. The \emph{Controller} is the person who is taking on the role of the robot. Due to the pandemic, we used Zoom to record the audio and create transcripts of the dialogue. We implemented a web-based data collection tool where the \emph{Controller} controls the agent in AI2-THOR through the web interface, and the \emph{Assistant} has access to a 2D scatter plot of a subset of objects in the scene (Fig.~\ref{fig:timeline}). Each pair of participants are assigned three object search trials in one environment. They have 90 seconds to explore the environment (with target objects removed) and 180 seconds to complete each trial. In addition to dialogue audio and transcripts, we collected data about the scene per view, the action executed, and the agent's groundtruth pose as provided by the AI2-THOR framework. We considered a discrete action space of \{\emph{MoveAhead}(0.25m), \emph{RotateLeft}(45$\degree$), \emph{RotateRight}(45$\degree$), \emph{LookUp}(30$\degree$), \emph{LookDown}(30$\degree$), \emph{Open}, \emph{Close}\}.\footnote{We first experimented with a rotation angle of 90$\degree$ following \cite{gordon2018iqa,ye2021hierarchical}, but experienced sudden jumps that are unnatural as felt by the participants. Therefore, we switch to 45$\degree$, also used by some existing works \cite{wortsman2019learning,qiu2020learning}}.

Despite the small scale of our pilot dataset, we observed some interesting behaviors shared between trials. For example, at the beginning of the object search trials the \emph{Assistant} would specify the target object and the \emph{Controller} would confirm. Additionally, as the task progresses, both roles would describe behaviors, beliefs about the environment and location of objects, and visual observations. We codified these into a set of preliminary intent types; some examples are given in the figure above. Using this pilot dataset, we have started to explore the development of an autonomous agent (\emph{Controller}), both modular and end-to-end that can plan actions for this task.

As mentioned in the introduction, we experimented with both speech-based dialogue and text-based dialogue, using the recording and chat features of Zoom. With speech, participants typically engage in frequent back-and-forth, as the \emph{Controller} controls the agent. Such exchanges involve discussing, for example, the scene and possible target locations. Participants report that when using text, the \emph{Controller} must decide between controlling the agent in AI2-THOR versus typing in the chat. Consequently, they would try to search for the object themselves without interacting with the \emph{Assistant}, who, as a result, finds it difficult to tell if their input is being considered by the \emph{Controller}. This suggests collecting dialogue data through text is unnatural and misaligned with our goal.

\section{Discussion \& Next Steps}
Though truthful to the task, our pilot data collection procedure is currently not scalable. We plan to implement a system that can be deployed on the crowdsourcing platform Amazon Mechanical Turk (AMT), to pair up Turkers to participate in the task entirely through their web browsers for accessibility. AMT a powerful platform, yet not designed for multi-user tasks. Due to audio communication and running AI2-THOR servers, we face a more difficult situation than ~\citet{das2017visual} who had to implemented a live chatbot on AMT. We also need solutions to scalable and accurate transcription of the collected audio as well as intent labling. We seek suggestions for strategies to collect such data at scale. In terms of evaluation, we believe both experiment with simulated assistants and real human assistants are necessary. For the simulated assistant, we are considering an oracle agent that communicates using template-based language. The goal of this simulated agent is to facilitate efficient and repeatable evaluations during algorithm development for the embodied dialogue agent, which could be a long-term effort. Ultimately, the agent should be deployed to perform the task with real human subjects. We plan to consider objective metrics for both object search performance (e.g., success rate and discounted total return\footnote{Because we consider open/close actions, the SPL metric \cite{batra2020objectnav} widely used in the object-goal navigation task is not applicable.}) and dialogue quality~\cite{venkatesh2018evaluating}, and, eventually, subjective metrics such as naturalness~\cite{hung2009towards}. We believe finding solutions to scalable speech-based dialogue data collection for embodied tasks and plausible evaluation protocol are daunting, yet unavoidable challenges towards future collaborative robots.

\bibliography{references}

\end{document}